\documentclass[letterpaper, 10 pt, conference]{ieeeconf}
\IEEEoverridecommandlockouts
\usepackage{amsmath,amssymb,amsfonts}
\usepackage{algorithmic}
\usepackage{textcomp}
\usepackage{xcolor}
\usepackage{algorithm}
\usepackage{array}
\usepackage[caption=false,font=normalsize,labelfont=sf,textfont=sf]{subfig}
\usepackage{stfloats}
\usepackage{url}
\usepackage{verbatim}
\usepackage{times}
\usepackage{epsfig}
\usepackage{float}
\usepackage{wrapfig}
\usepackage{bm,xspace}
\usepackage{comment}
\usepackage{multirow}
\usepackage{booktabs}
\usepackage{threeparttable}
\usepackage{graphicx}
\usepackage{balance}
\usepackage{etoolbox,siunitx}
\usepackage{calc}
\usepackage{pifont,hologo}
\usepackage{nicefrac}
\usepackage{makecell}
\usepackage{svg}
\usepackage{enumerate}
\makeatletter
\let\NAT@parse\undefined
\makeatother
\usepackage[colorlinks=black,citecolor=green]{hyperref}
\usepackage{adjustbox}
\usepackage{multicol}

\def\BibTeX{{\rm B\kern-.05em{\sc i\kern-.025em b}\kern-.08em
    T\kern-.1667em\lower.7ex\hbox{E}\kern-.125emX}}

\title{\LARGE \bf OpenBench: A New Benchmark and Baseline for Semantic Navigation in Smart Logistics}

\author{Junhui Wang\textsuperscript{1,2}, Dongjie Huo\textsuperscript{3}, Zehui Xu\textsuperscript{4}, Yongliang Shi\textsuperscript{2}, Yimin Yan\textsuperscript{5}, Yuanxin Wang\textsuperscript{6}, \\ Chao Gao\textsuperscript{2$\dag$}, Yan Qiao\textsuperscript{1$\dag$}, Guyue Zhou\textsuperscript{2,7$\dag$}  
\thanks{$^{1}$Institute of Systems Engineering and Collaborative Laboratory for Intelligent Science and Systems, Macau University of Science and Technology, $^{2}$Institute for AI Industry Research (AIR), Tsinghua University, $^{3}$College of Information Science and Technology, Beijing University of Chemical Technology, $^{4}$School of Astronautics, Harbin Institute of Technology, $^{5}$School of Artificial Intelligence, University of Chinese Academy of Sciences, $^{6}$School of Mechanical and Vehicular Engineering, Beijing Institute of Technology, $^{7}$School of Vehicle and Mobility, Tsinghua University.}%
\thanks{$\dag$ Corresponding authors: Chao Gao, Yan Qiao and Guyue Zhou.}%
\thanks{Sponsored by Xinchen Qihang Inc. and XIAOMI Fund.}
}

\begin{document}
\maketitle
\thispagestyle{empty}
\pagestyle{empty}

\begin{abstract}
The increasing demand for efficient last-mile delivery in smart logistics underscores the role of autonomous robots in enhancing operational efficiency and reducing costs. Traditional navigation methods, which depend on high-precision maps, are resource-intensive, while learning-based approaches often struggle with generalization in real-world scenarios. To address these challenges, this work proposes the Openstreetmap-enhanced oPen-air sEmantic Navigation (OPEN) system that combines foundation models with classic algorithms for scalable outdoor navigation. The system uses off-the-shelf OpenStreetMap (OSM) for flexible map representation, thereby eliminating the need for extensive pre-mapping efforts. It also employs Large Language Models (LLMs) to comprehend delivery instructions and Vision-Language Models (VLMs) for global localization, map updates, and house number recognition. To compensate the limitations of existing benchmarks that are inadequate for assessing last-mile delivery, this work introduces a new benchmark specifically designed for outdoor navigation in residential areas, reflecting the real-world challenges faced by autonomous delivery systems. Extensive experiments in simulated and real-world environments demonstrate the proposed system's efficacy in enhancing navigation efficiency and reliability. To facilitate further research, our code and benchmark are publicly available\footnote{\href{https://ei-nav.github.io/OpenBench/}{https://ei-nav.github.io/OpenBench/}\label{OpenBench}}.
\end{abstract}

\section{INTRODUCTION}
In the context of smart logistics, the demand for efficient and autonomous last-mile delivery is increasing rapidly. Autonomous robots offer a promising solution to meet this need, as they can enhance efficiency, improve customer experience, reduce costs, and minimize reliance on manual labor \cite{Liu2023}. To achieve these goals, navigation systems should be interactive, easy to deploy, and highly efficient. Traditional navigation methods requiring high-precision semantic maps are resource-intensive and hinder large-scale deployment \cite{Skog2009}. Meanwhile, recent learning-based approaches often struggle with generalization in real-world scenarios and require extensive training data \cite{shah2022gnm, shah2023vint, sridhar2023nomad}. To overcome these challenges, an Openstreetmap-enhanced oPen-air sEmantic Navigation (OPEN) system is proposed, combining foundation models with classic algorithms. This innovative approach provides a scalable solution for outdoor semantic navigation.

\begin{figure}[!t]
\centering
\includegraphics[width=0.48\textwidth]{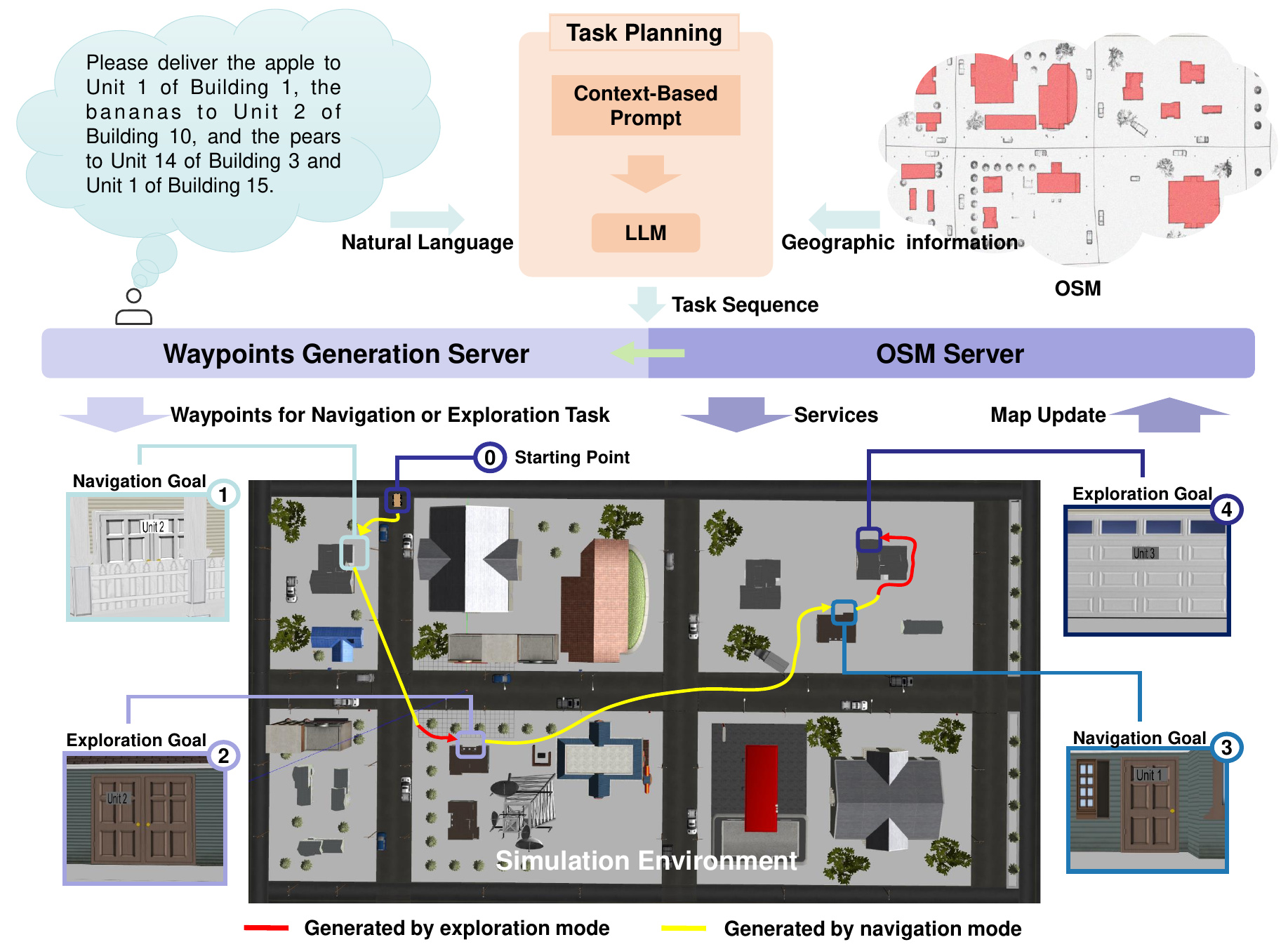}%
\caption{Overview of the proposed benchmark framework. The diagram presents the simulation environments and corresponding OSM, which are provided for the implementation of semantic navigation systems. This framework necessitates the navigation system to process natural language instructions autonomously, enabling accurate navigation from the initial starting point to the designated customer’s front door.}
\label{benchmark_framework}
\vspace{-4mm}
\end{figure}

The proposed approach uses OpenStreetMap (OSM) \cite{VargasMunoz2021} as a lightweight and flexible map representation, eliminating the need for labor-intensive pre-mapping. By using OSM, an open and crowdsourced resource, the OPEN system can dynamically interpret and navigate diverse outdoor environments. Furthermore, foundation models are incorporated to enhance system capabilities. Large Language Models (LLMs) are employed for the comprehension and analysis of delivery instructions, while Vision-Language Models (VLMs) facilitate global localization, map update, and house number recognition in open world. This ensures navigation stability, even in the absence of GPS.

Existing benchmarks for semantic navigation are observed to be primarily focused on indoor environments, making them insufficient for the requirements of large-scale outdoor semantic navigation in smart logistics \cite{Zhu_2021_CVPR,cows,Xia2020}. These benchmarks do not adequately capture the interactiveness and the long-term operational demands of real-world last-mile delivery. To address this gap, a new benchmark tailored for last-mile delivery in residential areas. This benchmark simulates real-world conditions, requiring robots to navigate from a designated starting point to a customer’s front door using only a navigation map, reflecting the practical challenges faced by human couriers.

In summary, the proposed method enhances the interactive and easy-to-deploy capabilities of robotic navigation systems by integrating foundation models and OSM. Moreover, the proposed benchmark compensates for the shortcomings of existing benchmarks by introducing evaluation metrics that specifically account for long-term operation in last-mile delivery. This combined effort is essential for driving forward the development of efficient, reliable, and scalable robotic delivery solutions in residential areas. The key contributions are as follows.

\begin{enumerate}
\item \textbf{New Benchmark for Last-Mile Delivery}. A new benchmark is introduced to optimize last-mile delivery in residential environments. It offers a framework for evaluating outdoor semantic navigation systems, focusing on long-term operational capability and task comprehension ability during delivery.

\item \textbf{Baseline Implementation}. The OPEN system is presented as a baseline for last-mile delivery in residential areas. It is interactive, easy to deploy, and uses off-the-shelf OSM for lightweight map representation, eliminating the need for pre-mapping.

\item \textbf{Combination of Foundation Models and Classic Algorithms}. The OPEN system combines foundation models and classic algorithms to enhance semantic navigation. It employs LLMs for natural language understanding and VLMs for global localization, map updates, and house number recognition. This approach ensures reliable GPS-free navigation, improving the system's efficiency, reliability, and long-term performance.

\item \textbf{Simulated and Real-World Experiments}. Extensive experiments in simulated and real-world environments validate the OPEN system's effectiveness in last-mile delivery. The results show significant improvements in navigation efficiency and reliability. To benefit the community, we make our code and benchmark accessible to the public\textsuperscript{\ref{OpenBench}}.
\end{enumerate}

\section{Related Work}
Achieving reliable and efficient navigation in autonomous mobile robots remains a significant challenge. Traditional navigation approaches, such as Simultaneous Localization and Mapping \cite{Cadena2016}, path planning \cite{SnchezIbez2021}, and robot control \cite{Tzafestas2018}, rely heavily on pre-constructed high-precision maps \cite{Skog2009}, limiting large-scale deployment, especially in last-mile delivery scenarios in residential areas.

Recent advances in learning-based navigation techniques, particularly reinforcement learning \cite{Hao2024,Liang2020}, offer promising alternatives by mapping sensory inputs directly to actions. Although promising, these approaches are predominantly tailored for short-range navigation and are constrained by the reality gap associated with on-policy reinforcement learning. NoMaD and ViNT \cite{sridhar2023nomad,shah2023vint} use goal images and topological graphs to facilitate visually guided robotic navigation. MTG and TGS \cite{Liang2024,tgs2024} employ a CVAE-based trajectory generation method to produce diverse candidate trajectories, subsequently selecting the most optimal one. Nevertheless, these learning-based methods often necessitate extensive training datasets and significant computational resources, and they frequently exhibit limited generalization capabilities across varying environments.

The advent of LLMs and VLMs has positioned semantic navigation as a promising direction for robotics \cite{VLA2024,Wang2024}. Gadre et al. \cite{cows} explore the use of the CLIP \cite{pmlr-v139-radford21a} model for language-driven zero-shot object navigation without additional training. Huang et al. \cite{Huang2023} introduce VLMaps, integrating pretrained visual-language features with 3D reconstructions to enable complex language-driven navigation. Yokoyama et al. \cite{Yokoyama2024} present Vision-Language Frontier Maps, combining occupancy maps with VLMs to achieve navigation in both simulated and real-world environments. While most research focuses on indoor navigation, Dhruv et al. \cite{shah2023lm} address the less-explored domain of outdoor semantic navigation, enabling complex tasks from natural language instructions without fine-tuning or annotated data.

To further advance the application and evaluation of navigation systems in outdoor environments, particularly for the last-mile delivery challenge in smart logistics, this paper proposes a corresponding benchmark and baseline. The proposed approach combines the strengths of traditional methods with those of foundation models, offering a robust solution for real-world scenarios.

\section{Last-Mile Delivery Benchmark}
The aim of this benchmark is to enhance the interactivity, ease of deployment, and long-term reliability of navigation systems in last-mile delivery contexts. As illustrated in the Fig. \ref{benchmark_framework}, users are encouraged to implement semantic navigation systems capable of processing delivery instructions and completing the corresponding tasks.
\subsection{Task Definition}
The last-mile delivery task involves interpreting natural language instructions and navigating autonomously from a starting point to a customer's residence. The system guides the robot to the customer's front door without the use of pre-constructed maps, relying instead on publicly accessible OSM navigation data. This setup closely mimics the real-world conditions encountered by human delivery personnel.

\begin{figure}[!t]
\centering
\includegraphics[width=0.48\textwidth]{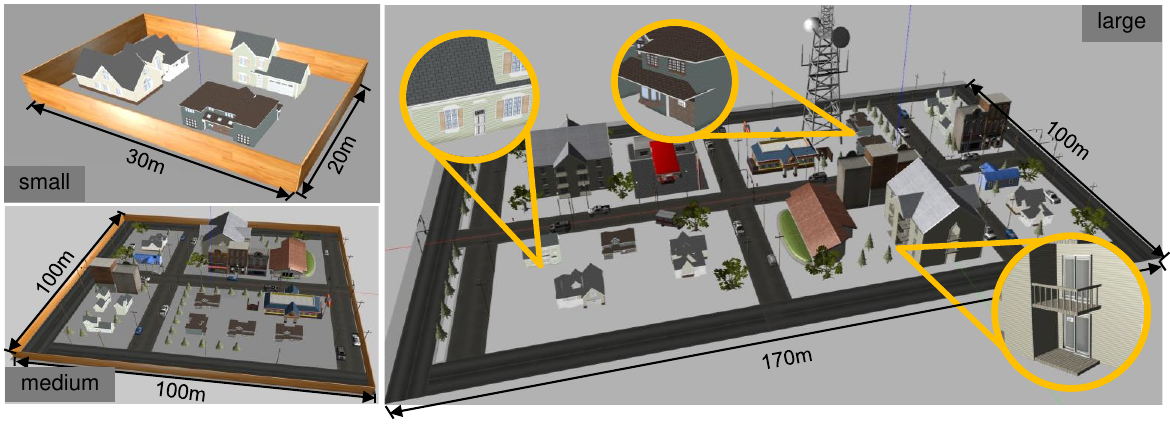}%
\caption{Simulation environment for last-mile delivery.}
\label{sim_env}
\vspace{-4mm}
\end{figure}

\subsection{Simulation Environment}
Based on gazebo simulation platform, we constructed
three distinct world models of varying sizes, categorized into three levels: small, medium, and large, depending on the
complexity of their environments. Each building within these
models has been labeled with house numbers on their doors,
as illustrated in Fig. \ref{sim_env}. Additionally, corresponding OSMs data are generated for each world model, reflecting real-world situations.

\subsection{Evaluation Metrics} \label{sec:eva}
To evaluate interactivity and long-term navigation in last-mile delivery tasks, we propose metrics to assess critical aspects of the process.

Success Rate of Task Planning (SRTP) quantifies the precision of task planning based on LLMs, reflecting the ability to understand instructions. It is defined as

\begin{equation}
SRTP = \frac{1}{N} \sum_{i=1}^{N} T_i
\end{equation}
where $N$ is the total number of delivery tasks, and $T_i$ is a binary variable indicating task success (1) or failure (0).

The overall success and efficiency of task completion are assessed using the Success Rate (SR) and Success Weighted by Path Length (SPL) metrics \cite{on2018}. SR represents the proportion of successfully completed tasks, while SPL incorporates both task completion and path efficiency.

For sequential delivery, maintaining consistent performance over time is critical. Since delivery tasks are tightly interconnected following task planning, the failure of earlier tasks can adversely affect the execution of subsequent ones. For instance, if each task has a specific deadline, the failure of preceding tasks will impact the completion times of all subsequent tasks. To evaluate this, the \textbf{Long-term Success Rate (LSR)} is introduced, extending the SR metric by considering task success across continuous operations.

\begin{equation}
LSR = \frac{1}{N} \frac{\sum_{i=1}^{N} c_i \cdot S_i}{\sum_{i=1}^{N} c_i}
\end{equation}
where $S_i$ indicates the success of task $i$, and $c_i$ is a weighting factor derived from an exponential decay model.

The \textbf{Long-term Success Weighted by Path Length (LSPL)} metric further refines this evaluation by considering both task success and navigation efficiency over time. LSPL is defined as

\begin{equation}
LSPL = \frac{1}{N} \frac{\sum_{i=1}^{N} c_i \cdot S_i \cdot \frac{l_i}{\max(p_i, l_i)}}{\sum_{i=1}^{N} c_i}
\end{equation}
where $l_i$ represents the shortest path distance from the starting point to the goal in task $i$, while $p_i$ is the actual path length taken.

The weighting factor $c_i$ in both LSR and LSPL follows an exponential decay model.

\begin{equation}\label{ci}
c_i = \frac{r^{i-1} \cdot (1 - r)}{1 - r^n}, \quad i = 1, 2, 3, \dots, n
\end{equation}
where $r$ is the decay rate, $i$ denotes the task sequence, and $n$ is the total number of tasks. This model prioritizes earlier tasks, recognizing their influence on subsequent deliveries.

The LSR and LSPL metrics provide comprehensive insights into long-term performance, complementing established SR and SPL metrics to evaluate the sustained efficiency of robotic delivery solutions.

\begin{figure*}[!t]
\centering
\includegraphics[width=1.0\textwidth]{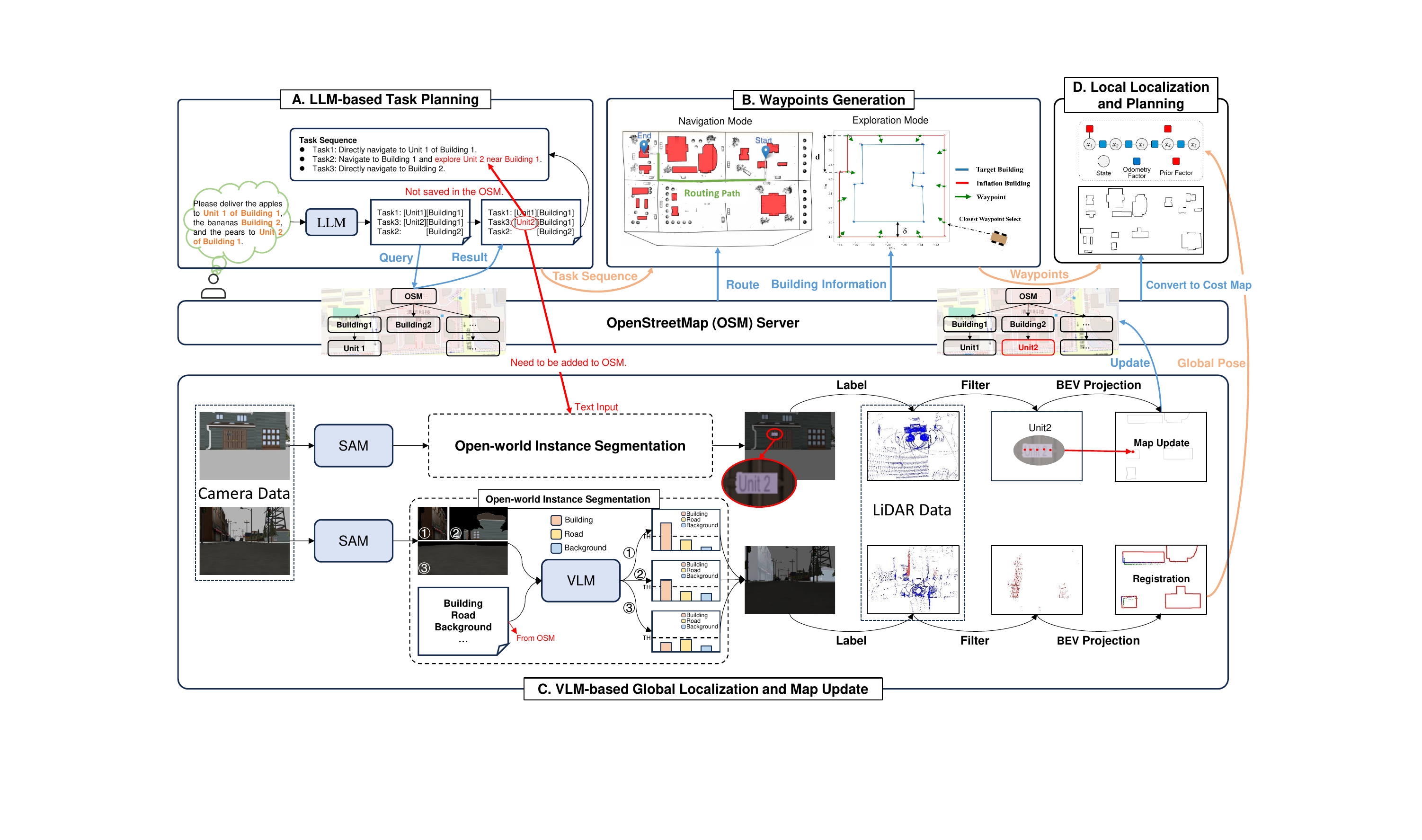}%
\caption{Overview of the OPEN system for autonomous last-mile delivery. The system initiates with a natural language delivery request, processed by a task planning module powered by an LLM. This module interacts with OSM to extract destination details and generates a structured task sequence. The robot autonomously decides between navigation and exploration modes, generating waypoints for execution by a classical planner. Localization is performed using classical methods, with global localization enhanced through integration of MobileSAM and CLIP models with OSM to correct positional errors. The robot also updates OSM with newly detected objects, continuously improving map detail and navigation performance for subsequent deliveries.}
\label{navigation_baseline}
\vspace{-4mm}
\end{figure*}

\section{Proposed Method}
\subsection{System Overview}
The OPEN system for autonomous last-mile delivery is illustrated in Fig. \ref{navigation_baseline}. The process begins with a natural language delivery request, processed by a task planning module based on an LLM. This module interacts with OSM \cite{VargasMunoz2021} to extract destination details and generate a structured task sequence. The robot then autonomously determines whether to operate in navigation or exploration mode, generating waypoints for execution by a classical planner.

For local pose estimation, the robot uses classical localization methods. To mitigate cumulative localization errors and align with the OSM coordinate system, the system performs global localization at lower frequencies by integrating MobileSAM \cite{2023computer} and CLIP \cite{pmlr-v139-radford21a} models with OSM. The robot also detects and identifies objects in its environment, updating OSM to enhance map detail and improve navigation for future deliveries.

By combining LLMs and VLMs, the system offers a robust solution for last-mile delivery without pre-mapping. It is interactive, easy to deploy, and highly efficient, providing accurate navigation while continuously enriching OSM. This ongoing map enhancement improves long-term navigational performance and adaptability to real-world scenarios.

\subsection{LLM-based Task Planning}
This work presents a task planning approach that utilizes LLMs to convert multilingual, free-form text instructions into structured robotic tasks. The method is divided into three key phases, illustrated in Fig. \ref{navigation_baseline}A.
\subsubsection{Address Resolution} Users provide delivery instructions in natural language, often containing multiple tasks. The initial LLM prompt extracts and parses a series of addresses, subdividing them into hierarchical sub-addresses. To mitigate "hallucinations" (incorrect outputs generated by LLMs), a secondary prompt is employed to verify the extracted information. Prompting the LLM to emulate human-like reasoning further reduces the risk of errors.
\subsubsection{Task Optimization} After address extraction, another prompt optimizes the task sequence. Geographically proximate tasks are grouped for simultaneous completion, while tasks across regions are modeled as a classical scheduling problem to enhance efficiency and reducing time.
\subsubsection{Location Query} The method queries the hierarchical sub-addresses within OSM from the lowest to the highest levels. The query process halts once an address is confirmed to exist at a certain level. The query outcomes fall into two categories: (i) if the lowest level sub-address is present in OSM, the robot navigates directly based on this information, and (ii) if any sub-addresses are absent from OSM, the robot first navigates to the lowest known sub-address and subsequently explores for the missing lower-level sub-addresses. Thus, the task sequence required for each delivery is generated based on the completeness of OSM information.

\subsection{Waypoints Generation}
Upon receiving task sequences, robots execute navigation or exploration activities as guided by OSM to complete deliveries. This work introduces two waypoint generation modes for navigation and exploration, as shown in Fig. \ref{navigation_baseline}B.
\subsubsection{Navigation Mode}
For the higher-level components of the delivery address, OSM generally includes positional information, allowing us to utilize OSM's road network data to generate global routing guidance \cite{luxen-vetter-2011}. Leveraging OSM for routing requires the following two steps.
\begin{itemize}
    \item \textbf{Road Preprocessing}: OSM data is converted into a hierarchical graph based on a transportation-specific profile (e.g., vehicle, pedestrian) for efficient routing.
    \item \textbf{Route Querying}: The routing process utilizes the Multi-Level Dijkstra (MLD) algorithm. MLD reduces the search space by exploiting the hierarchical graph, allowing for the rapid determination of optimal paths between geographic coordinates.
\end{itemize}

This approach offers a scalable solution for generating accurate and efficient routing across diverse geographic regions and transportation modes.

\subsubsection{Exploration Mode}
When OSM lacks fine-grained details, such as precise building entrances, the robot autonomously explores to locate the entrance through the following steps.
\begin{itemize}
    \item \textbf{Building Preprocessing}: The robot begins by retrieving the target building’s location and geometry from OSM. To obtain the building’s outer boundary and exclude internal elements, the robot calculates the concave hull of the building. This polygon is then inflated to ensure a reliable search area around the building’s perimeter.
    \item \textbf{Uniform Sampling}: The inflated polygon is uniformly sampled, generating waypoints oriented towards the building’s centroid to optimize search coverage and guide the robot to potential entrances.
    \item \textbf{House Number Recognition}: At each waypoint, the robot captures an image via its RGB camera, which is analyzed by a VLM to identify the target entrance. If the VLM determines that the target entrance has not been located, the robot advances to the next waypoint, continuing this process until the target entrance is successfully detected.
\end{itemize}


\subsection{VLM-based Global Localization and Map Update}
Enhancing autonomous navigation in complex environments requires robust global localization. We propose a novel method leveraging VLMs and OSM for localization, aligning the robot’s position within OSM coordinates, reducing odometry drift, and providing a reliable GPS alternative in urban areas. It also supports dynamic map updates by incorporating new elements, improving adaptability in real-world scenarios.
\subsubsection{Global Localization} To address odometry errors and unreliable GPS in urban settings, this work uses VLMs for direct localization with OSM. As shown in Fig. \ref{navigation_baseline}C, the MobileSAM model \cite{2023computer} segments objects in images, while OSM provides element types it contains (e.g., buildings, roads). Segmented images and OSM text are encoded into a shared embedding space using CLIP \cite{pmlr-v139-radford21a}, and the system assigns semantic labels to objects based on calculated probabilities. Labeled results are aggregated to identify OSM elements within the segmented images. A point cloud is then projected onto the images for semantic information, and relevant points are retained and projected onto a Bird’s Eye View (BEV) plane. Finally, 2D registration with OSM-derived geometry provides the robot’s global pose.
\subsubsection{Map update} For last-mile delivery tasks, online map updates enhance the robot’s memory. This process mirrors global localization, with the key difference being the addition of newly detected elements, such as house number plates, to the map. By adding more granular address information into OSM, the system improves the efficiency of future deliveries.

A key advantage of this method is its flexibility in element types for both localization and map updates. By leveraging CLIP's zero-shot generalization capabilities, the system can adapt to open-world environments, similar to human cognition, without being limited to predefined elements.

\subsection{Local Localization and Planning}
Accurate local state estimation and efficient path planning are essential for robotic navigation, as shown in Fig. \ref{navigation_baseline}D. We use FastLIO2 \cite{Xu2021} for precise LiDAR-based local pose estimation, maintaining the transformation between the LiDAR and odometry frames. Concurrently, a factor graph \cite{Kaess2011} integrates local state estimations as odometry factors and global estimations as prior factors. Upon successful global localization, a new prior factor is added and the graph is optimized, updating the transformation between the map and odometry frames. This method ensures bounded localization error over time, even without GPS.
For path planning, OSM is converted into a costmap for A* pathfinding, followed by the Timed Elastic Band algorithm \cite{Rsmann2017} to generate smooth, dynamically feasible trajectories for real-time control.

\begin{figure}[!t]
\centering
\includegraphics[width=0.46\textwidth]{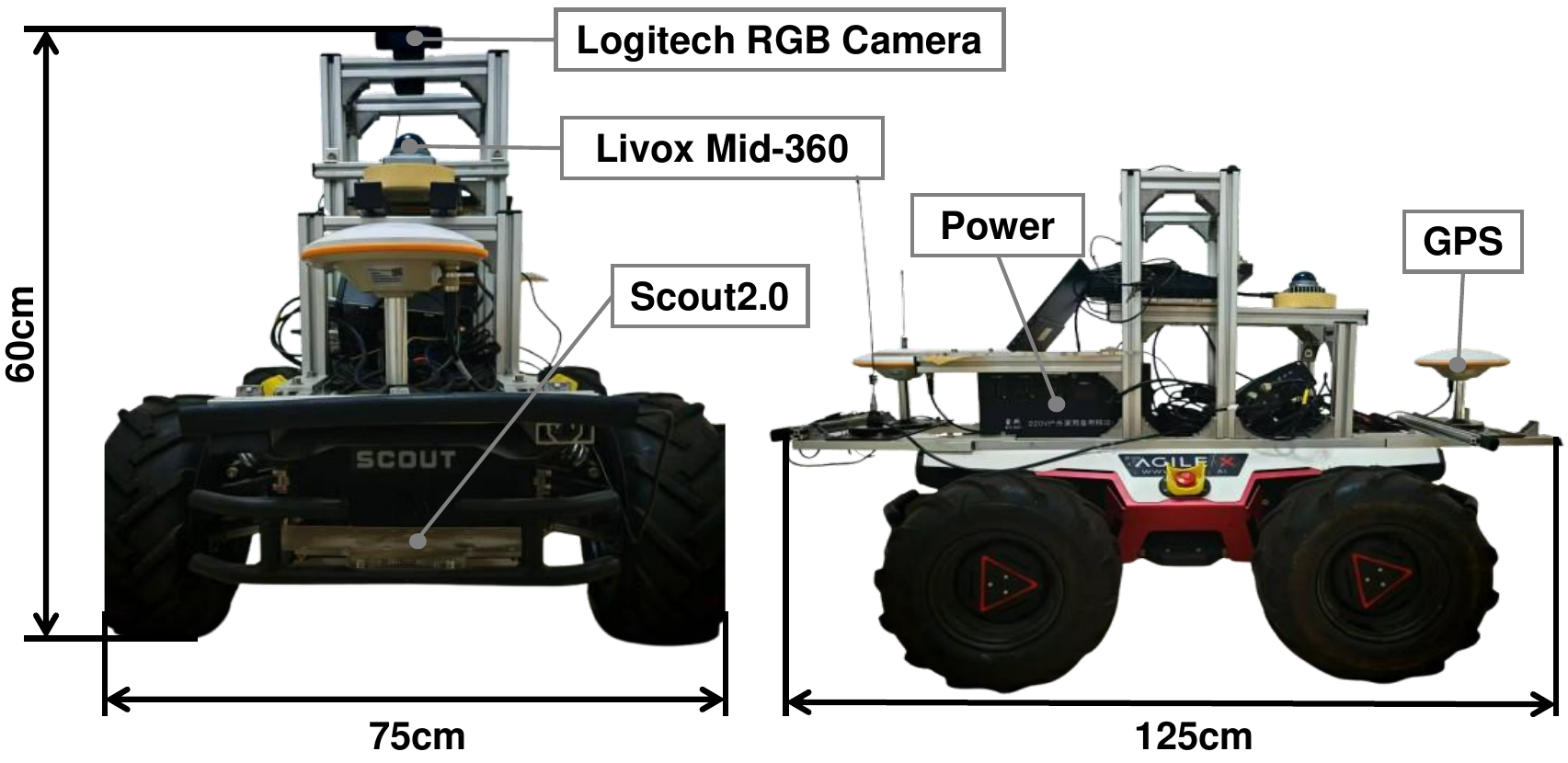}%
\caption{The robot used in real-world navigation experiments.}
\label{exper_platform}
\vspace{-4mm}
\end{figure}

\begin{table*}[t]
\caption{Comparative analysis of navigation performance across various semantic navigation methods in simulation environment.} 
\centering
\label{nav_perf}
\begin{tabular}{@{}cccccccccccccccc@{}}
\toprule
\multirow{2}{*}{Method} &  & \multicolumn{4}{c}{Small} &  & \multicolumn{4}{c}{Medium} &  & \multicolumn{4}{c}{Large} \\ \cmidrule(lr){3-6} \cmidrule(lr){8-11} \cmidrule(l){13-16} 
                        &  & SR  & SPL  & LSR  & LSPL  &  & SR   & SPL  & LSR  & LSPL  &  & SR  & SPL  & LSR  & LSPL  \\ \midrule
ViNT~\cite{shah2023vint}                    &  & 40$\%$   & 21.13$\%$    & -    & -     &  & 20$\%$    & 17.67$\%$    & -    & -   &  & 0   & 0    & -    & -     \\
NoMaD~\cite{sridhar2023nomad}                   &  & 40$\%$   & 9.31$\%$    & -    & -    &  & 20$\%$    & 18.23$\%$    & -    & -  &   & 0   & 0    & -    & -     \\
OPEN                    &  & \textbf{100$\%$}   & \textbf{35.93$\%$}    & \textbf{100$\%$}    & \textbf{61.31$\%$}  &   & \textbf{100$\%$}    & \textbf{32.24$\%$}    & \textbf{75.58$\%$}    & \textbf{12.92$\%$}  &   & \textbf{60$\%$}   & \textbf{31.12$\%$}    & \textbf{83.98$\%$}    & \textbf{47.97$\%$}    \\ \bottomrule
\end{tabular}
\begin{tablenotes}
    \item The symbol "-" indicates that the evaluation is not applicable.
\end{tablenotes} 
\end{table*}

\section{EXPERIMENTS}
\subsection{Experiment Setup}
\subsubsection{Simulation Environment} As shown in Fig. \ref{sim_env}, the simulation uses a differential-drive, four-wheel robot equipped with a monocular camera and a Livox MID-360 LiDAR. A state recorder tracks task completion and movement trajectory, providing data for evaluation metrics.
\subsubsection{Real-world Environment} In the real-world setup (Fig. \ref{exper_platform}), the robot is equipped with a Logitech RGB camera, a Livox MID-360 LiDAR, and a GPS. The GPS records the robot's actual trajectory, while task completion is assessed manually. The manually controlled path serves as the optimal trajectory for comparison.
\subsubsection{Evaluation Metrics}
We use the metrics from Section \ref{sec:eva}, setting \( r = 0.9 \) in Eq. \ref{ci}. Successful navigation is defined as the robot reaching within 10m of the destination.
\subsubsection{Computing Platform}
The benchmark and baseline tests are conducted on a PC with an AMD R9-7945HX processor and an RTX 4060 GPU.

\subsection{Results on Simulation Environment}
In the simulation experiments, delivery destinations are randomly generated and organized into text instructions (e.g. Fig.\ref{navigation_baseline}A) or goal images as input for the navigation system.
\subsubsection{Sucess Rate of Task Planning} The success of task planning depends on LLMs' ability to accurately parse destination addresses into a specified format (JSON in this work). Failures in planning affect subsequent navigation tasks, emphasizing the importance of task comprehension. We evaluate several LLMs, each undergoing 60 test runs. As shown in Table \ref{task_planning}, GPT-4O-mini achieved the highest success rate, while others show room for improvement.

\begin{table}[!t]
\caption{Comparative analysis of success rate of task planning}
\centering
\label{task_planning}
\begin{tabular}{@{}cccc@{}}
\toprule
\makebox[0.12\textwidth][c]{LLM} & \makebox[0.07\textwidth][c]{SRTP} & \makebox[0.12\textwidth][c]{LLM} & \makebox[0.07\textwidth][c]{SRTP} \\ \midrule
Gemini-1.5-pro   & 0.27  & GPT-3.5-turbo  & 0.9    \\
Qwen-turbo       & 0.5   & Claude-3.5     & 0.97   \\
WenXinYiYan      & 0.6   & GPT-4o-mini    & 1.0    \\
\bottomrule
\end{tabular}
\end{table}

\subsubsection{Navigation Performance} To our knowledge, no open-source methods align with our objectives. Therefore, the proposed method is compared with NoMaD and ViNT \cite{sridhar2023nomad,shah2023vint}, which are learning-based navigation systems that use goal images and topological graphs to enable mobile robots to navigate toward goals. They require pre-collected images to construct the topological map.
To evaluate SR and SPL, five individual tasks are tested with system restarts between trials. For LSP and LSPL, continuous delivery to five destinations is assessed. NoMaD and ViNT are not evaluated for LSP and LSPL due to their lack of multi-task capability.

As summarized in Table \ref{nav_perf}, both NoMaD and ViNT demonstrate poor generalization in simulation environments, with low success rates primarily due to collisions during navigation. Their success is limited to simple tasks. In contrast, the proposed OPEN system, which leverages OSM guidance and the VLM module, achieved a higher success rate, outperforming both NoMaD and ViNT. In the LSR and LSPL evaluations, the proposed method completes five, three, and three tasks in small, medium, and large simulation environments, respectively. Compared to executing tasks individually, the number of successful tasks decreased, indicating that sequential execution impacts performance. However, due to the higher weight of earlier tasks, the system achieved a higher score in the large simulation environment despite completing the same number of tasks. Overall, the system maintains strong performance in long-term navigation.

\subsubsection{Influence of Map Update}
We assess the impact of map updates by conducting navigation to three randomly generated destinations, both with and without known door locations. SPL is calculated to quantify efficiency improvement, which increased by 30.61$\%$, 6.31$\%$, and 47.87$\%$ for the three destinations, as shown in Table \ref{map_update}.

\begin{table}[!t]
\caption{The impact of map updates on navigation efficiency}
\centering
\label{map_update}
\begin{tabular}{@{}cccc@{}}
\toprule
     & \makebox[0.11\textwidth][c]{[unit1][building7]} & \makebox[0.11\textwidth][c]{[unit2][building16]} & \makebox[0.11\textwidth][c]{[unit2][building12]} \\ \midrule
OPEN w/o & 22.44$\%$ & 85.68$\%$ & 34.72$\%$ \\
OPEN w/ & 29.31$\%$ & 91.09$\%$ & 51.34$\%$  \\ \bottomrule
\end{tabular}
\begin{tablenotes}
    \item w/o indicates without map updates, and w/ indicates with map updates.
\end{tablenotes}
\end{table}

\subsubsection{Map Storage Efficiency Evaluation}
A lightweight map representation is essential for practical robotic applications. We compare OSM and point cloud maps with the topological map used in NoMaD. The point cloud map is downsampled using a 0.2m voxel grid. As shown in Table \ref{map_stor}, our map's storage space is approximately 1$\%$ of the point cloud and 0.01$\%$ of the topological map, demonstrating its efficiency.
\begin{table}[!t]
\caption{Comparison of storage space}
\centering
\label{map_stor}
\begin{tabular}{@{}cccc@{}}
\toprule
  \makebox[0.1\textwidth][c]{} & \makebox[0.04\textwidth][c]{OSM} & \makebox[0.1\textwidth][c]{Point cloud map} & \makebox[0.14\textwidth][c]{Topological map \cite{sridhar2023nomad}} \\ \midrule
Small  & \textbf{5.2kB}   & 499.7kB               & 70MB               \\
Medium & \textbf{26.9kB}   & 5.9MB               & 315.62MB               \\
Large  & \textbf{37.1kB}   & 10.2MB     & 720.15MB           \\ \bottomrule
\end{tabular}
\end{table}

\begin{figure}[!t]
\centering
\includegraphics[width=0.48\textwidth]{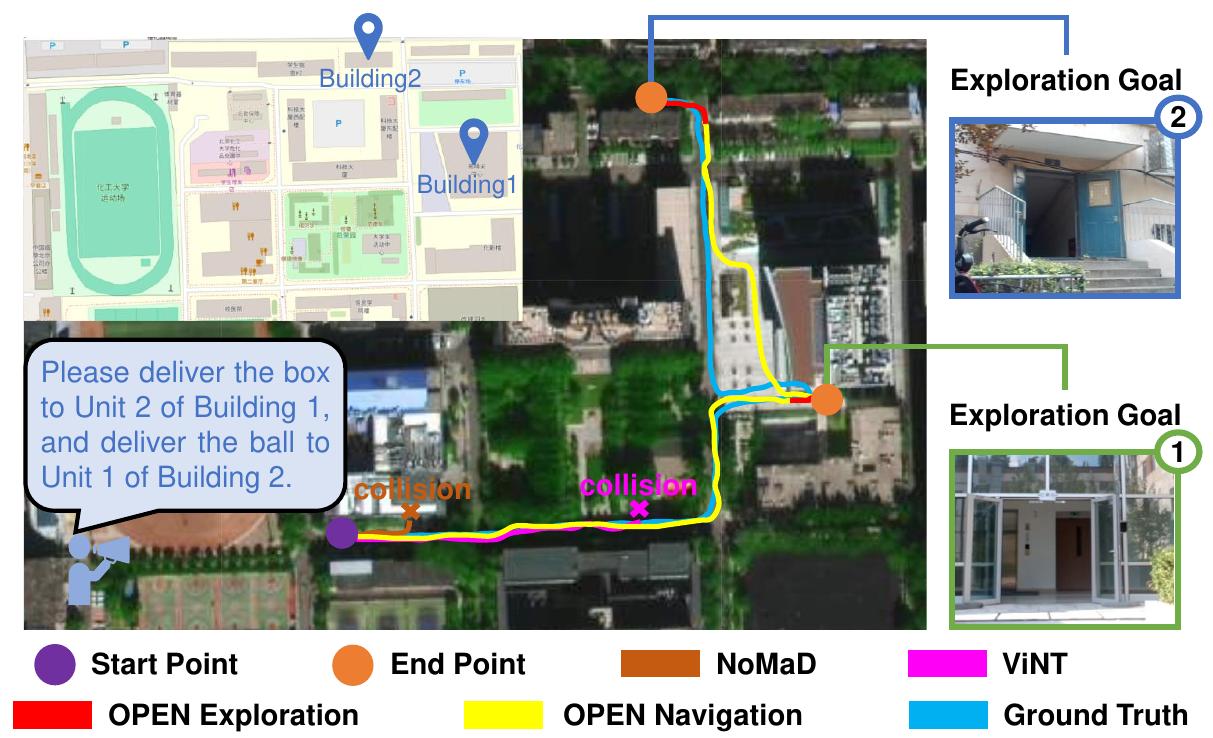}%
\caption{Illustration of the real-world experiment. The top-left part presents the OSM and target buildings. The bottom-left part displays the delivery instructions. The right side of the figure shows the navigation trajectories of different methods.}
\label{real_nav}
\vspace{-4mm}
\end{figure}

\subsection{Results on Real-world Environment}
The experiments are conducted in a real-world campus environment, utilizing OSM data for the Beijing University of Chemical Technology (BUCT) area, which were directly obtained from the OpenStreetMap website, as illustrated in Fig. \ref{real_nav}. The experimental task involves sequential deliveries to two distinct buildings. Four different approaches are evaluated: ViNT, NoMaD, the proposed OPEN system, and a human-operated remote control baseline. The navigation trajectories for each method are depicted in Fig. \ref{real_nav}.

Both the ViNT and NoMaD systems encounter collisions during navigation to the first building, ultimately failing to complete the task. In contrast, the OPEN system successfully executes the entire delivery sequence, demonstrating performance closely aligned with that of the human-operated control, which achieves an SPL of 96.1$\%$.

\section{CONCLUSIONS}
In conclusion, this work introduces the OPEN system, a novel approach to last-mile delivery that integrates OSM with advanced foundation models to tackle the challenges of scalable and efficient outdoor navigation. By using OSM for lightweight map representation and incorporating LLMs and VLMs for global localization, map update, and house number recognition, the system overcomes the limitations of traditional map-based and learning-based methods. The introduction of a new benchmark, specifically designed for last-mile delivery, offers an effective framework for evaluating autonomous delivery systems. Extensive experiments in both simulated and real-world environments show significant improvements in navigation efficiency, reliability, and long-term operational capability. This demonstrates the system's potential for direct deployment in various residential settings without the need for pre-mapping.

\bibliographystyle{IEEEtran}  
\bibliography{ref} 
\end{document}